\documentclass{article}[10pt]

\usepackage{config}

\title{\bf On the Properties of Adversarially-Trained CNNs}
\author{Mattia Carletti, Matteo Terzi, Gian Antonio Susto\\
University of Padova\\
{\tt \small mattia.carletti@studenti.unipd.it, \{matteo.terzi,gianantonio.susto\}@unipd.it}}
\date{}

\begin{document}

\maketitle

\begin{abstract}
Adversarial Training has proved to be an effective training paradigm to enforce robustness against adversarial examples in modern neural network architectures. Despite many efforts, explanations of the foundational principles underpinning the effectiveness of Adversarial Training are limited and far from being widely accepted by the Deep Learning community. In this paper, we describe surprising properties of adversarially-trained models, shedding light on mechanisms through which robustness against adversarial attacks is implemented. Moreover, we highlight limitations and failure modes affecting these models that were not discussed by prior works. We conduct extensive analyses on a wide range of architectures and datasets, performing a deep comparison between robust and natural models.
\end{abstract}

\section{Introduction}
\label{sec:intro}
Deep Neural Network (DNN) architectures have enjoyed huge success in image classification tasks \cite{lecun2015deep, krizhevsky2009learning, krizhevsky2012imagenet}. However, since the discovery of adversarial examples \cite{szegedy2013intriguing, goodfellow2014explaining}, the trustworthiness of their predictions has began to be questioned. To date, many defenses have been proposed to ameliorate the robustness against adversarial attacks, but only few of them proved to be still effective after improvement of existing attacks \cite{athalye2018obfuscated, tramer2020adaptive}. Arguably the most prominent amongst such successful defenses, Adversarial Training (AT) \cite{madry2017towards} has become a cornerstone and the main benchmark for robustness in DNNs. As such, countless efforts have been made towards understanding the principles underlying the tremendous success of AT. Nonetheless, this long-standing problem is far from being solved. In this work, we build on well-established behaviors of adversarially-trained models and provide a thorough comparison with respect to natural models. Our analysis goes beyond known properties and limitations of robust models\footnote{The expressions \emph{robust models} and \emph{adversarially-trained models} will be used interchangeably hereinafter.} - such as the accuracy-robustness trade-off \cite{tsipras2018robustness, yang2020closer} or the shape bias \cite{zhang2019interpreting} - and unveil previously unnoticed behaviors. Moreover, we provide experiments aiming at linking hypotheses with recent findings in related literature.
The aim of this work is twofold. Firstly, to reveal properties of adversarially-trained models that could serve as a base for future research with the objective of achieving robustness without AT. Secondly, to challenge some common beliefs on robust models as a first step towards establishing more sound empirical evidence of their limitations. 
The focus of this paper is on Convolutional Neural Network (CNN) architectures commonly adopted in image classification tasks. Based on the results of our experiments, we conclude that current CNNs are not the best choice for the goal of robust classification in Computer Vision. Our contributions can be summarized as follows:
\begin{itemize}
    \item We show that feature maps in adversarially-trained models are more dense and activate more frequently (i.e., for more data points in the dataset) compared to natural models.
    \item We show that feature maps in adversarially-trained models are more redundant than in natural models, thus reducing the effective number of channels in hidden layers.
    \item We show that the latent space of adversarially-trained models offers representations with different degrees of robustness. This confirms the fact that AT preserves extra information about the input that is ignored by the robust classifier.
    \item We assess the color bias of adversarially-trained models and point out subtle failure modes that may undermine the deployment of robust models in practice.
\end{itemize}
We first overview prior works on AT and biases of CNNs in \Cref{sec:related_work}, then introduce the methodological tools needed for our analyses in \Cref{sec:methods}. Experimental results are presented in \Cref{sec:exp} and thoroughly discussed in \Cref{sec:discussion}. In \Cref{sec:conclusion} we sum up our findings and draw the conclusions.

\section{Related Work}
\label{sec:related_work}
\paragraph{Adversarial Training}
It is widely accepted that AT represents the current state-of-the-art in defending against adversarial attacks. Indeed, while many other defenses have been proposed \cite{guo2017countering,dhillon2018stochastic,xie2017mitigating, song2017pixeldefend}, they have been sistematically evaded by refined attacks \cite{athalye2018obfuscated, tramer2020adaptive}. Being the de-facto standard for enforcing robustness, AT has attracted a great deal of research interest. Many studies analyzed structural properties \cite{ilyas2019adversarial, engstrom2019adversarial} and peculiar behaviors \cite{salman2020adversarially, terzi2020adversarial, utrera2020adversarially, yang2020closer} of adversarially-trained models. 
In particular, a well-known problem affecting robust models is the so-called accuracy-robustness trade-off \cite{tsipras2018robustness, yang2020closer}. Initially, this tension was thought to be inherent when training robust models \cite{tsipras2018robustness}, but successive analysis \cite{yang2020closer} proved that commonly used image datasets are actually separable, conjecturing that a perfectly robust and accurate classifier can, in principle, exist. Other studies focused on improving the performance of AT.
In \cite{xie2020adversarial}, the authors propose to use distinct batch norm layers for clean and adversarial examples during training to improve natural accuracy.
In \cite{zhang2019theoretically} a novel formulation of adversarial defense, dubbed TRADES, is proposed. The method is based on the optimization of a loss taking into account both natural accuracy and the adversarial robustness.
Finally, another line of research is devoted to the analysis and estimation of the Lipschitz constant of DNNs, as a guarantee of stability and robustness to be enforced during training \cite{scaman2018lipschitz, huang2021training, liang2020large}.
The present work complements prior studies on the properties of robust models and provide arguments to resolve contrasting results found in the literature.
\paragraph{Biases of CNNs}
In \cite{ding2019sensitivity}, the authors analyze the impact of semantics-preserving transformations of the input data distributions on clean accuracy and adversarial robustness. The experiments show that robust accuracy under Projected Gradient Descent training is much more sensitive than clean accuracy under standard training to the differences in input data distribution. In \cite{de2021impact}, the impact on natural accuracy of color distortions is assessed. Specifically, the authors propose a variant of the ImageNet dataset where a set of color distortions is applied to original images.
The aim of \cite{hermann2020origins} is to investigate the source of the texture bias in models trained on ImageNet. It shows that random-crop augmentation biases the models towards texture and proposes more naturalistic forms of data-augmentation as a simple way to mitigate texture bias. Surprisingly, they could extract and decode with high accuracy shape information from hidden layers. This suggests that classification layers might play an important role in removing shape information. 
In \cite{zhang2019interpreting}, the authors exploit saliency maps to interpret the inner workings of adversarially-trained models and compare them to standard models. They also evaluate these models on distorted test sets preserving either shape or textures and verify that adversarially-trained models rely more on global features such as shape and edges. They finally show that standard models are biased towards textures, as previously observed by \cite{geirhos2018imagenet}. 
Along these lines, our experiments focus on the simplicity biases of robust models. Different from previous studies, primarily focused on the bias towards shape and textures, our analysis is addressed to the bias towards color.

\section{Methods}
\label{sec:methods}
Throughout the paper we consider classifiers encoded by CNNs of the form $f \circ g$, where $f$ is a feature extractor with learnable parameters ${\boldsymbol{\theta}_f}$ and $g$ is a fully-connected linear layer with learnable parameters ${\boldsymbol{\theta}_g}$. We refer to the outputs of $f$ as \emph{latent features}. CNNs considered in this work are composed of stacked sequences of convolutional layers (\texttt{Conv}), Batch Normalization layers (\texttt{BN}), Rectified Linear Unit (\texttt{ReLU}) activation functions and pooling layers.
Natural models are trained by optimizing a suitable loss function $\mathcal{L}$ on image-label pairs $(\mathbf{x}, y)$ drawn from the training set $\mathcal{D}$. In our experiments, we consider the cross-entropy loss and Stochastic Gradient Descent (SGD) \cite{bottou2018optimization} as optimizer.
Robust models are trained according to the Adversarial Training (AT) protocol with PGD  first-order adversary \cite{madry2017towards}. Specifically, during training the parameters of the model are optimized so as to minimize the loss on adversarial examples rather than on original examples. 
In this work, we consider PGD-based adversarial attacks. Given a norm $\| \cdot \|_{p}$ and a perturbation budget $\epsilon>0$, an adversarial example $\mathbf{x}'$ for the original image $\mathbf{x}$ (with ground truth label $y$) must satisfy $\|\mathbf{x}' -\mathbf{x}\|_p \leq \epsilon$. If we denote with $B_{\epsilon}$ the $\ell_p$-ball of radius $\epsilon$ centered at $\mathbf{x}$, the adversarial example is initialized at a random point in $B_{\epsilon}$ and iteratively updated (for a given number of steps) according to the following rule:
\begin{align}\label{eq:adv_ex}
\mathbf{x}'_{k+1} &= \Pi_{B_{\epsilon}}(\mathbf{x}'_k + \alpha \cdot \mathbf{g}) \\
\mathbf{g} &= \argmax_{\|\mathbf{u}\|_p \leq 1} \mathbf{u}^\top \nabla_{\mathbf{x}'_k} \mathcal{L}(\mathbf{x}'_k, y). \nonumber
\end{align}
In \Cref{eq:adv_ex}, $\alpha$ is the attack step-size, $\Pi_{B_{\epsilon}}(\cdot)$ projects an input onto $B_{\epsilon}$, and $\mathbf{g}$ is the gradient, that represents the steepest ascent direction for a given $\ell_p$-norm.
We denote with $\epsilon_{tr}$ the perturbation budget used during the training and with $\epsilon_{te}$ the perturbation budget to craft adversarial examples for robustness evaluation. The robust models analyzed in this paper are trained/evaluated on attacks bounded in either $\ell_2$- or $\ell_{\infty}$-norm.

Before describing the analytical tools employed in our experiments, we introduce the necessary notation.
Let partition the feature extractor $f$ of a generic CNN in \emph{blocks} along the depth dimension and denote with $\mathbf{x}^k$ the activations (corresponding to input image $\mathbf{x}$) at the output of the $k$-th block. The definition of \emph{block} depends on the architecture. By way of example, for models in the ResNet family \cite{he2016deep, zagoruyko2016wide} we collect activations at the output of each ResNet block, any of whom is composed of multiple \texttt{Conv}-\texttt{BN}-\texttt{ReLU} sequences.
In general, $\mathbf{x}^k$ is a 3-D tensor with dimension ${C_k \times H_k \times W_k}$, where $C_k$, $H_k$ and $W_k$ are the number of channels, feature maps height and width at the output of the $k$-th block. For our purposes, we consider its 2-D version $\Tilde{\mathbf{x}}^k \in \mathbb{R}^{C_k \times (H_k \times W_k)}$ with vectorized feature maps.
Let $\Tilde{\mathbf{x}}^k[i]$ (or equivalently, $\mathbf{x}^k[i]$) represent the $i$-th feature map at the output of the $k$-th block, with $i \in \{1, \dots, C_k \}$.

\begin{definition}
We say that a feature map is \emph{$\tau_{act}$-active} if at least one of its activations is greater than $\tau_{act}$, for a given activation threshold $\tau_{act} \in \mathbb{R}_+$. 
\end{definition}

\begin{definition}
We say that a feature map is \emph{densely $\tau_{act}$-active at level $\tau_{dens}$} if at least $\tau_{dens} \cdot 100 \; \%$ of its activations are greater than $\tau_{act}$, for a given activation threshold $\tau_{act} \in \mathbb{R}_+$ and density threshold $\tau_{dens} \in \left] 0, 1\right]$. 
\end{definition}

\begin{definition}
Let $\mathcal{I}^k_+$ be the subset of active feature maps indices at the output of the $k$-th block. We define the tensor of the active feature maps at the output of the $k$-th block as
$\Tilde{\mathbf{x}}^k_+ \defeq \Tilde{\mathbf{x}}^k[\mathcal{I}^k_+].$
\end{definition}
Notice that $\Tilde{\mathbf{x}}^k_+$ has dimension $|\mathcal{I}^k_+| \times H_k \times W_k$, where $|\mathcal{I}^k_+|$ is the cardinality of $\mathcal{I}^k_+$ and $|\mathcal{I}^k_+| \leq C_k$. For the sake of notation simplicity, we will not mention $\tau_{act}$ and $\tau_{dens}$ when their values are clear from the context.

\subsection{Measuring Densely Active Feature Maps}
\label{subsec:methods_alw_active}
\paragraph{Motivation}
Many works discussed the interplay between model sparsity and robustness to adversarial attacks \cite{xiao2018training, wong2018provable, guo2018sparse, ye2019adversarial, ozdenizci2021training, wang2018adversarial}, often leading to opposite conclusions \cite{guo2018sparse}. Since the parameters of a model provide a convenient static representation of the task, these studies are mostly focused on weight sparsity.  Differently, in this work we address the sparsity of activations. Indeed, the interaction between the input and the model - rather than a structural representation of the latter - is the key factor to be investigated to characterize the model behavior.
\paragraph{Methodology} Sparsity of activations can be assessed by counting, for each data point in the dataset and each block in the model, the number of densely active feature maps. We combine this information with the frequency of activation of densely active feature maps across the dataset to get a measure of expressivity of the model. 

\subsection{Measuring Feature Maps Redundancy}
\label{subsec:methods_redundancy}
\paragraph{Motivation}
We focus on an underexplored direction to understand how adversarially-trained CNNs work internally, namely the redundancy of feature maps. We draw inspiration from three facts: i) redundant signals are widely exploited in contexts where robustness to noise is a primary concern, e.g., the theory of frames \cite{kovacevic2008introduction} or error-correcting codes \cite{guruswami2008explicit}; ii) the trade-off between accuracy and robustness in DNNs is conjectured to be due to limitations of existing architectures and/or training methods \cite{yang2020closer}, rather than class separability under adversarial attacks \cite{tsipras2018robustness}; iii) AT benefits more from larger capacity than standard training \cite{madry2017towards, xie2019intriguing}.
The presence of higher internal redundancy in adversarially-trained models would provide a valid argument to justify empirical observations ii) and iii).
\paragraph{Methodology} Given a block index $k$, the starting point is the computation of the cosine similarity matrix $S^k_+$ between active feature maps
\begin{equation}
    S^k_+(i, j) = \frac{<\Tilde{\mathbf{x}}^k_+[i], \Tilde{\mathbf{x}}^k_+[j]>}{\| \Tilde{\mathbf{x}}^k_+[i] \| \cdot \| \Tilde{\mathbf{x}}^k_+[j] \|}
\end{equation}
where $<\cdot,\cdot>$ represents the inner product.
The next step consists in clustering together active feature maps whose cosine similarity is above a given threshold $\tau_{sim}$. As a result, we obtain a set of clusters $\mathcal{C}^k_1, \dots, \mathcal{C}^k_{n_k}$, with $n_k \leq C_k$. Completely uncorrelated feature maps would produce $n_k=C_k$ clusters with one element each, while high correlation ($> \tau_{sim}$) amongst some of the feature maps would result in a more limited number of clusters, some of whom would have cardinality $>1$. 
The number of clusters at the output of a given block can be thought of as the \emph{effective number of active channels} at the output of the block.

\begin{definition}
We say that an active feature map is \emph{redundant} if it belongs to a cluster with cardinality $>1$.
\end{definition}
According to the definitions given above, for the $k$-th block we analyze feature maps redundancy by means of the following two quantities:
    (i) the number of redundant feature maps $C_k^{R}$;
    (ii) the average value of cosine similarity $\bar{S}^k_+$, computed over the off-diagonal elements of $S^k_+$.
Notice that the number of redundant feature maps gives an intuitive and direct measure of redundancy in the activations, though it does not capture correlations below the threshold $\tau_{sim}$. The average value of cosine similarity takes this into account and complements in this sense the information provided by $C_k^{R}$. We compute the average cosine similarity instead of the Frobenius norm of $S^k_+$ since the latter depends on the number of active feature maps, which is variable for different input images.
\subsection{Inherent Robustness of Latent Features}
\label{subsec:methods_latent}
\paragraph{Motivation}
A recent work \cite{shafahi2019adversarially} demonstrated that robustness can be preserved in transfer learning settings, where the feature extractor - trained on the source domain - is kept freezed and the linear classifier is retrained on natural examples from the target domain. The authors also showed that retraining the linear classifier on natural examples from the source domain does not affect the robust accuracy. However, this behavior has been assessed for a single model and whether this claim can be generalized should be investigated. This is the goal of our experiments, where we analyze the predictive power and robustness of latent representations in adversarially-trained models. We draw inspiration from the surprising fact that representations in robust models can be inverted \cite{engstrom2019adversarial}. This property might appear incompatible with the accuracy-robustness trade-off observed in practice: if the representation is invertible, then information about the data should be preserved in the latent space and natural accuracy of robust models should not drop. In \cite{terzi2020adversarial}, the authors resolve such discrepancy by proving that AT preserves information about the data, but the information that is \emph{accessible} to the classifier does not contain all the details about the input. Our results corroborate this finding and show that latent representations of adversarially-trained models offer different degrees of robustness, dependent on the features combinations selected by the classifier.
\paragraph{Methodology} The setup for this batch of experiments is as follows. We consider a model $f \circ g$ pretrained in adversarial settings. We keep the feature extractor $f$ fixed and retrain from scratch the linear classifier $g$ \emph{on natural examples}. In other words, we re-initialize and then optimize ${\boldsymbol{\theta}_g}$, while ${\boldsymbol{\theta}_f}$ is kept unchanged. We find that this simple procedure can lead to non-trivial improvements in natural accuracy for robust models, but this occurs at the price of a (even more) significant decrease in robustness.
\subsection{Assessing Color Bias}
\label{subsec:methods_color_bias}
\paragraph{Motivation}
It is a known fact that adversarially-trained models are more biased towards simple features - such as shape and color - rather than textures \cite{utrera2020adversarially, chen2020shape}. While these so-called \emph{simplicity biases} are typically welcomed since they help improving robustness, they could be harmful in other contexts (as we shall show in \Cref{subsec:exp_color_bias}). In our experiments, we study the color bias of adversarially-trained models, a property that has been hitherto overlooked and not properly analysed. Indeed, most of the studies in the literature have been focusing on the dichotomy texture-shape \cite{geirhos2018imagenet, utrera2020adversarially, chen2020shape}.
\paragraph{Methodology} We assess the color bias of adversarially-trained and natural models with two experiments:
\begin{enumerate}
    \item We measure the natural accuracy on pixel-averaged images. In other words, for each image in the test set we consider its texture-less and shape-less monochromatic counterpart, where each pixel assumes the average value of all pixels (examples in \Cref{fig:avg_images_imagenet}).
    \item We consider transformed images where a colored contour of varying thickness is applied. We measure the natural accuracy drop with respect to the evaluation on clean images. The colored contours can be red, green, blue or white. For the CIFAR-10 dataset, we consider contours of thickness 1, 2, 3 or 4 pixels. For the ImageNet dataset, we consider contours of thickness 1, 5, 10, 15, 20 pixels (examples in \Cref{fig:contour_images_imagenet}).
\end{enumerate}
Notice that the semantic content of images is minimally affected by the added contour - at least for small thickness values - and humans would not be induced to make wrong predictions because of the frame element. Unlike humans, the performance of adversarially-trained CNNs can be significantly impaired in this context. One may argue that colored contours look unrealistic, but similar scenarios could be encountered in real-world images. It is worth mentioning the natural framing technique in photography, where elements of a scene (often unrelated from the main subject of the composition) are leveraged to create a natural frame within the photograph. Beyond that, we tested the effect of colored contours just as a prime example of how a content-preserving alteration of the original image can lead to dramatic changes in the behavior of these models. Hence, we do not exclude the existence of even more subtle case studies (see \Cref{fig:grid_images_imagenet,fig:grid_exp_imagenet} in \Cref{app:omitted_results} for another example). In the light of our findings, we actually encourage further investigations along these lines.

\section{Experiments}
\label{sec:exp}
\paragraph{Models and datasets} We conduct our analyses on three datasets: CIFAR-10 \cite{krizhevsky2009cifar}, CIFAR-100 \cite{krizhevsky2009cifar} and ImageNet \cite{deng2009imagenet}. For CIFAR-10 and CIFAR-100, we consider \rn{18} models trained with $\epsilon_{tr} \in \{0, 0.5, 1, 2, 3, 4\}$ and $\ell_2$-norm. The case $\epsilon_{tr}=0$ represents standard training. For ImageNet, we consider \rn{18}, \rn{50} \cite{he2016deep}, \wrn{50}{2}, \wrn{50}{4} \cite{zagoruyko2016wide} models trained with $\epsilon_{tr} \in \{0, 0.5, 3, 5\}$ for $\ell_2$-norm and with $\epsilon_{tr} \in \{0, 0.5, 1, 2, 4, 8\}$ for $\ell_{\infty}$-norm. Moreover, we consider \vgg~\cite{simonyan2014very} models trained with $\epsilon_{tr} \in \{0, 3\}$ and $\ell_2$-norm. More details on the training of the considered models are provided in \Cref{app:models}. Some experiments are based on a subset of the available models for the ease of visualization and/or for computational efficiency. Results on robust models trained with $\ell_{\infty}$-norm are presented in \Cref{app:omitted_results}.

\subsection{Densely Active Feature Maps}
\label{subsec:exp_alw_active}
We set $\tau_{dens}=0.95$ - similar results hold for other large enough values of $\tau_{dens}$. By examining how the number of densely active feature maps varies under different training conditions, a remarkable result emerges: compared to natural models, robust models have a higher number of feature maps that are densely active for \emph{all samples in the dataset}. This holds true across \emph{all architectures and datasets} analyzed in this work, as shown in \Cref{fig:num_alw_l2}, where $N_k$ is the number of always densely active feature maps for the $k$-th block.
Notice that in most cases, the larger $\epsilon_{tr}$, the higher the value of $N_k$. This means that, as the level of robustness of the model increases, intermediate blocks make less use of non-linearities, thus reducing the complexity of functions that can be encoded: a larger value of $N_k$ would entail reduced expressivity of the model as ReLUs have no effect on activations that operate in their linear region for all samples.

\renewcommand{\fsize}{0.24}
\begin{figure*}
    \centering
\includegraphics[width=\fsize\textwidth]{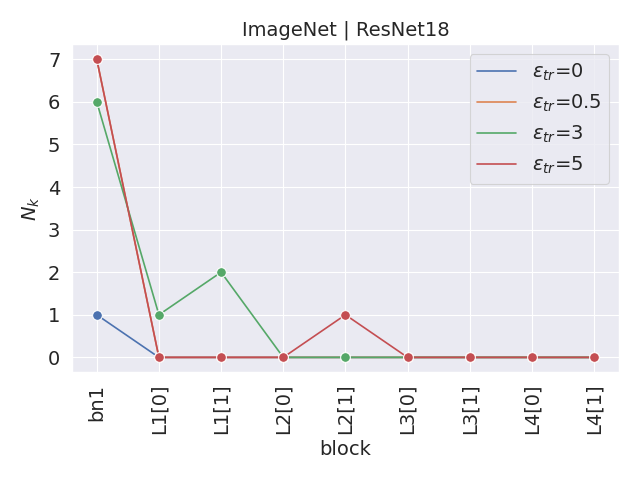}
\includegraphics[width=\fsize\textwidth]{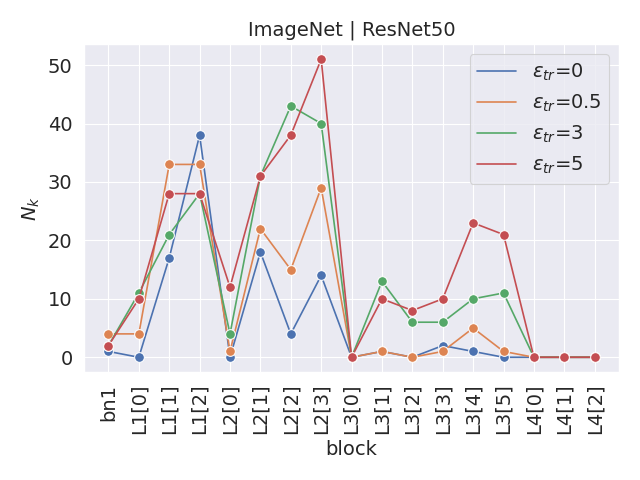}
\includegraphics[width=\fsize\textwidth]{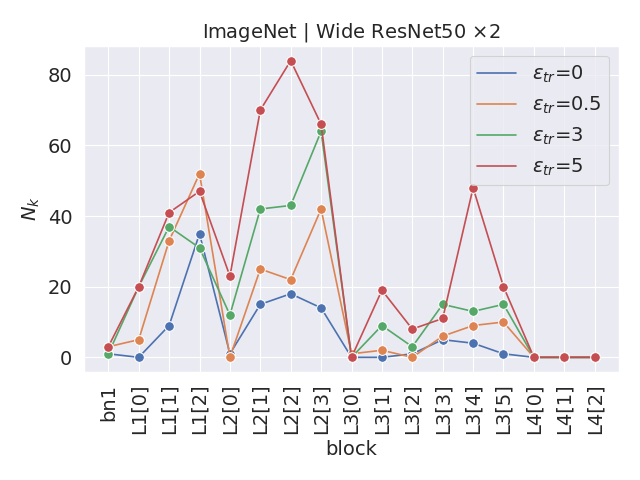}
\includegraphics[width=\fsize\textwidth]{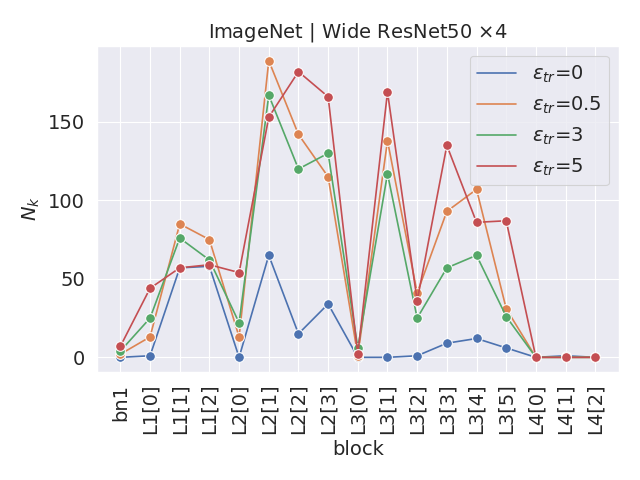}
\includegraphics[width=\fsize\textwidth]{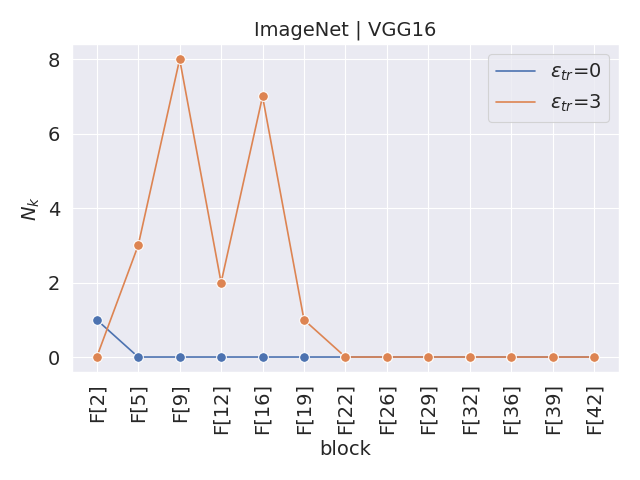}
\includegraphics[width=\fsize\textwidth]{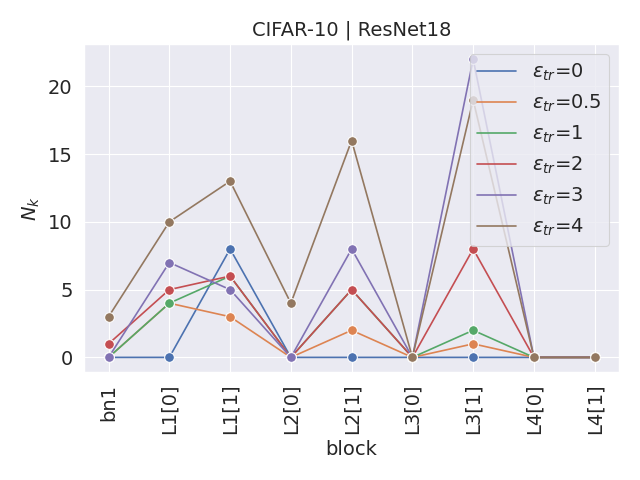}
\includegraphics[width=\fsize\textwidth]{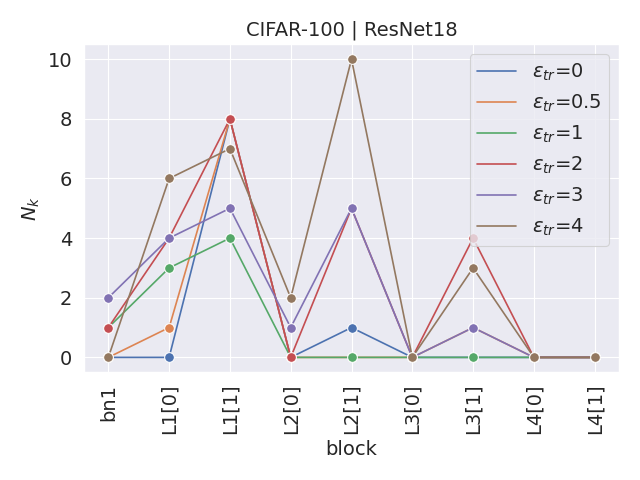}
    \caption{Number of always densely active feature maps (at level $\tau_{den}=0.95$). Robust models are trained with $\ell_2$-norm.}
    \label{fig:num_alw_l2}
\end{figure*}

The exposed experiments show that, due to AT, a new architecture structure emerges where there are two sets of feature maps: always active (or almost always active), dubbed $z_1$, and all the remaining ones, dubbed $z_2$. It is natural to ask what is the role of $z_1$.
By definition, always active activations must convey information that is always necessary/important. Thus, it is reasonable to think that $z_1$ controls in some way the flow of information of $z_2$.
Our hypothesis is that $z_1$ act as modulating (gating) factor with respect to $z_2$ such that holds the factorization $y \sim h_1(z_2 \mid z_1) h_2(z_1)$. 
One simple case could be $y = z_1 \odot h(z_2)$. Interestingly, this is very similar to the Gating Units proposed in \cite{liu2021pay}. Thus, we may suggest that AT creates a simple form of attention mechanism. We will explore in-depth the role of always active feature maps in future works.

\subsection{Feature Maps Redundancy}
\label{subsec:exp_redundancy}
In \Cref{fig:num_redundant_l2} are reported the number of redundant feature maps $C_k^R$ across layers for different architectures and training conditions (we set $\tau_{sim}=0.95$ to get clusters whose elements are highly correlated to each other). Notice that there exists a strong correlation between the level of robustness of adversarially-trained CNNs and the redundancy of their feature maps. The same conclusion can be drawn from the analysis of the average cosine similarity $\bar{S}^k_+$ in \Cref{fig:avg_sim_l2}, which is independent from the value of $\tau_{sim}$. This effect is most prominent for models with higher capacity. It is worth mentioning that wide natural models exhibit a larger number of redundant feature maps compared to their thinner counterparts.
This is consistent with the results in \cite{casper2019frivolous}, where the existence of so-called \emph{redundant units} is proved and leveraged to explain implicit regularization in wide (natural) models.
Our findings suggest a novel direction for investigation about the mechanism through which local robustness may be implemented by adversarially-trained CNNs, namely a coupling between feature maps. Notice that in \cite{liang2020large}, the authors propose - altough for a very simple DNN - a coupling between subsequent layers as a viable solution for achieving small Lipschitz constants, regardless of their norm. Results presented above suggest that AT might exploit similar schemes.

\renewcommand{\fsize}{0.24}
\begin{figure*}
    \centering
\includegraphics[width=\fsize\textwidth]{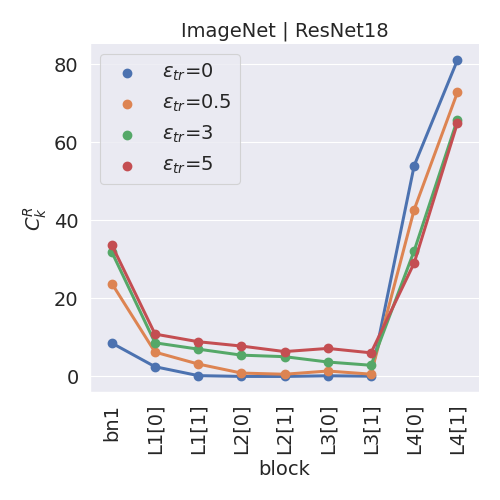}
\includegraphics[width=\fsize\textwidth]{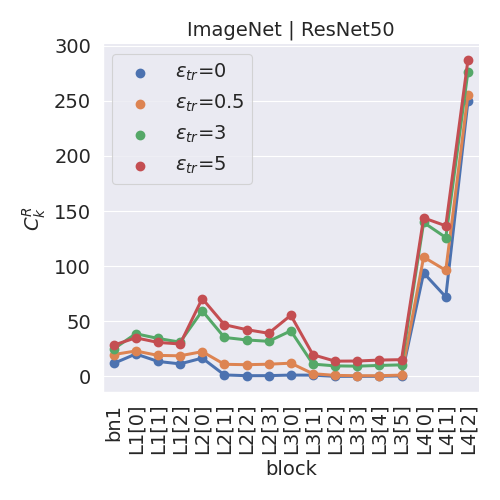}
\includegraphics[width=\fsize\textwidth]{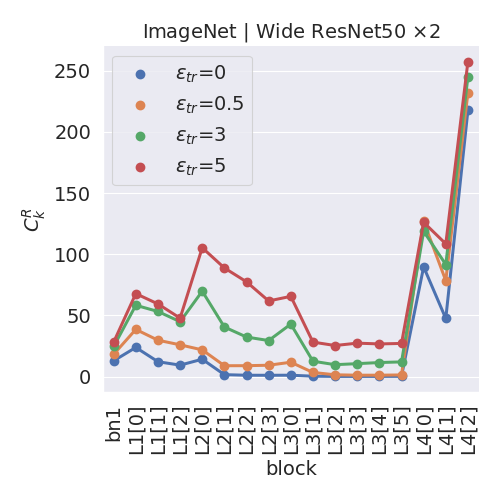}
\includegraphics[width=\fsize\textwidth]{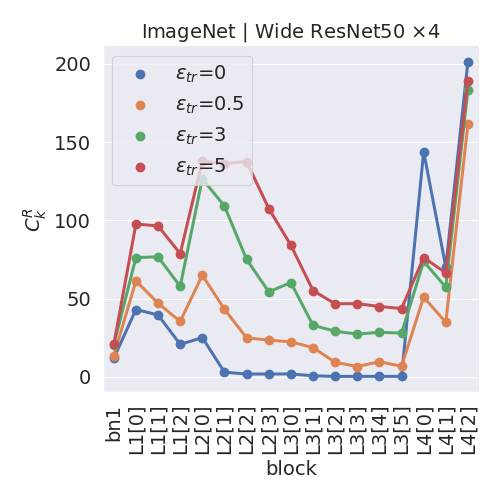}
\includegraphics[width=\fsize\textwidth]{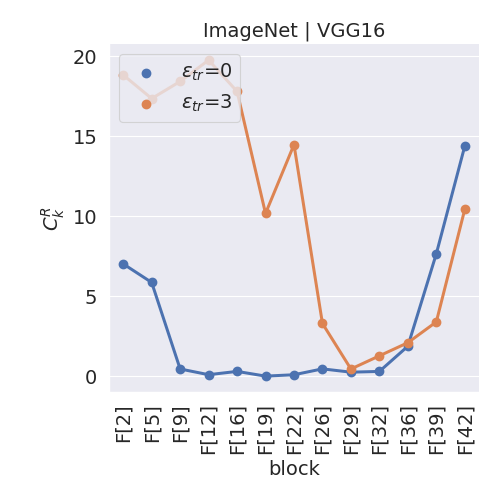}
\includegraphics[width=\fsize\textwidth]{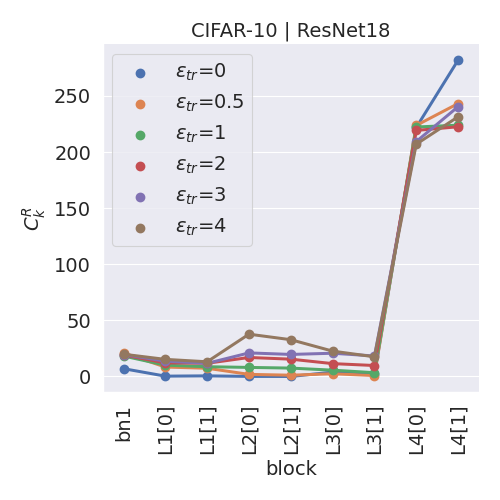}
\includegraphics[width=\fsize\textwidth]{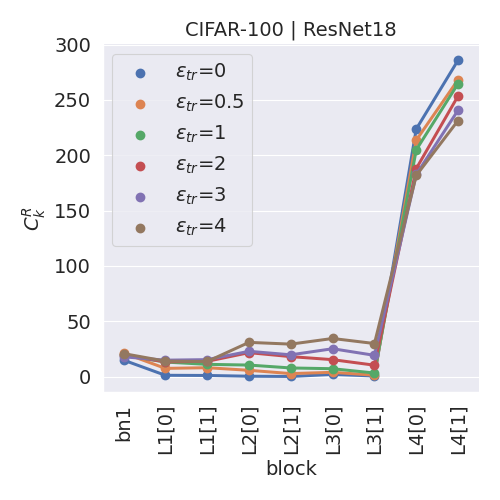}
    \caption{Number of redundant feature maps. Robust models are trained with $\ell_2$-norm. For ImageNet, Values are averaged over 5000 images randomly sampled from the test set.}
    \label{fig:num_redundant_l2}
\end{figure*}

\renewcommand{\fsize}{0.24}
\begin{figure*}
    \centering
\includegraphics[width=\fsize\textwidth]{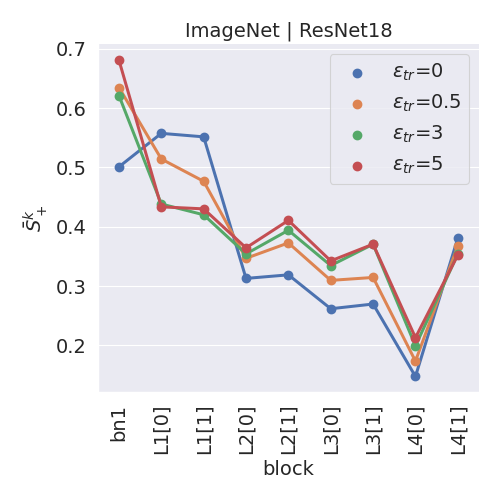}
\includegraphics[width=\fsize\textwidth]{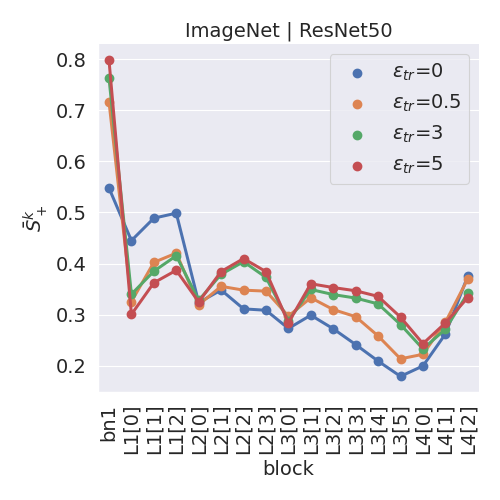}
\includegraphics[width=\fsize\textwidth]{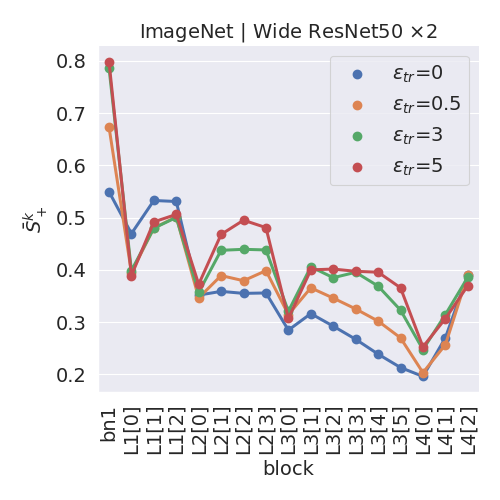}
\includegraphics[width=\fsize\textwidth]{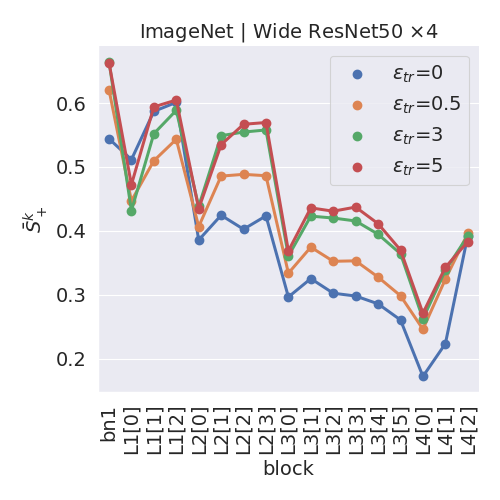}
\includegraphics[width=\fsize\textwidth]{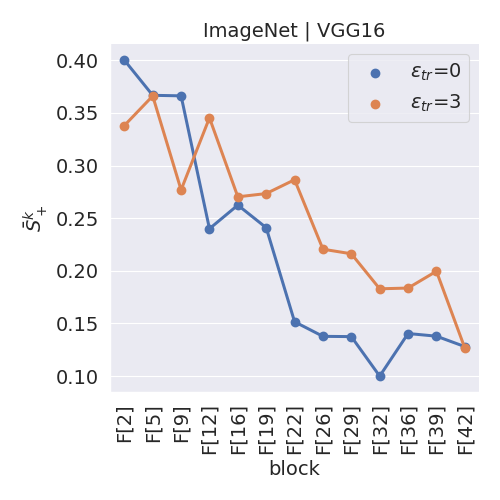}
\includegraphics[width=\fsize\textwidth]{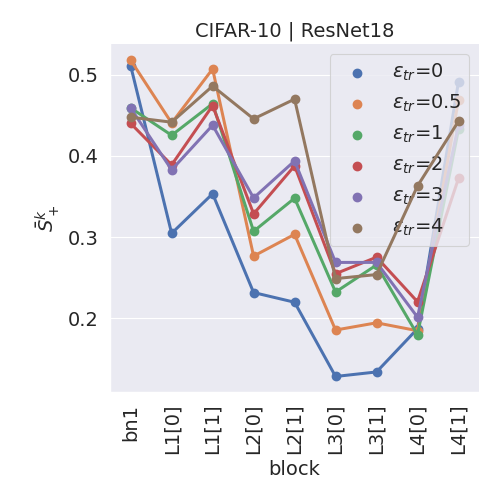}
\includegraphics[width=\fsize\textwidth]{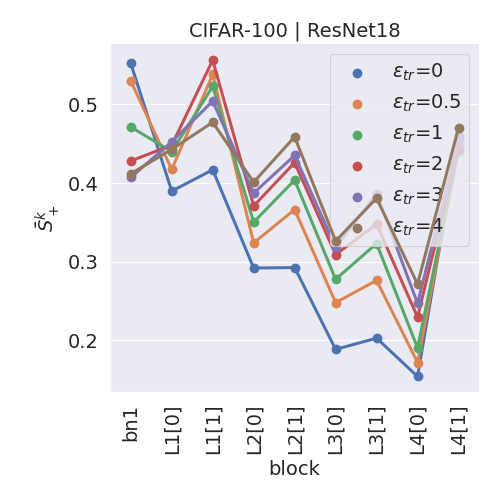}
    \caption{Average cosine similarity. Robust models are trained with $\ell_2$-norm. For ImageNet, Values are averaged over 5000 images randomly sampled from the test set.}
    \label{fig:avg_sim_l2}
\end{figure*}

\subsection{Latent Features}
\label{subsec:exp_latent}
We retrain linear classifiers $\tilde{g}$ for $15$ epochs on natural examples from the source domain on which the pretrained model $f \circ g$ was trained. Results are reported in \Cref{tab:clf_cifar} and \Cref{tab:clf_imagenet} for CIFAR-10/CIFAR-100 and ImageNet, respectively.
This procedure leads to a clear improvement in natural accuracy with respect to the original classifier $g$, unveiling the availability of predictive latent representations that were not fully exploited. On the other hand, such enhanced predictive power implies a drastic drop in robustness. As anticipated in \Cref{subsec:methods_latent}, our results are in line with \cite{terzi2020adversarial}: the information about the input is preserved in the latent space, but it is not fully accessible for the classifier.
We acknowledge a correlation between the value of $\epsilon_{tr}$ and the entity of this behavior - the larger $\epsilon_{tr}$, the larger the increase (decrease) in natural (robust) accuracy. 
We deduce that it may be possible to create a model structure where features are sorted by their degree of robustness (similarly to what happens with PCA). This would enable, with a unique model, to switch seamlessly between tasks that require different levels of robustness.

\begin{table}
\caption{Natural and robust accuracy of retrained classifiers on CIFAR-10 and CIFAR-100. Feature extractors are pretrained robust models trained with $\ell_2$-norm. Robust accuracy is evaluated against adversarial attacks crafted with 20 PGD steps, attack step size 1 and $\epsilon_{te}=\epsilon_{tr}$.}
\label{tab:clf_cifar}
\vskip 0.15in
\begin{center}
\begin{footnotesize}
\begin{sc}
\begin{tabular}{lllrrrrr}
\toprule
model & metric & classifier & $\epsilon_{tr}=0.5$ &  $\epsilon_{tr}=1$ &  $\epsilon_{tr}=2$ & $\epsilon_{tr}=3$ & $\epsilon_{tr}=4$ \\
\midrule
CIFAR-10 \rn{18} & natural accuracy & original & 88.34 & 80.14 & 64.43 & 59.25 & 46.54 \\
        &  & retrained & 88.78   & 82.53   & 72.23  & 67.69   & 56.48 \\
        &  &           & (+0.44) & (+2.39) & (+7.8) & (+8.44) & (+9.94) \\
        & robust accuracy & original & 68.30 & 51.44 & 33.38 & 22.85 & 16.26 \\
        &  & retrained & 67.16   & 46.77   & 20.36    & 8.71     & 4.87 \\
        &  &           & (-1.14) & (-4.67) & (-13.02) & (-14.14) & (-11.39) \\
\midrule
CIFAR-100 \rn{18} & natural accuracy & original & 63.68 & 57.97 & 49.73 & 37.41 & 27.42 \\
        &  & retrained & 64.16   & 57.60   & 52.34   & 46.03   & 39.56 \\
        &  &           & (+0.48) & (-0.37) & (+2.61) & (+8.62) & (+12.14) \\
& robust accuracy & original & 36.07 & 22.50 & 13.07 & 9.54 & 6.35 \\
        &  & retrained & 36.67  & 21.44   & 8.16    & 4.05    & 1.67 \\
        &  &           & (+0.6) & (-1.06) & (-4.91) & (-5.49) & (-4.68) \\
\bottomrule
\end{tabular}
\end{sc}
\end{footnotesize}
\end{center}
\vskip -0.1in
\end{table}

\begin{table}
\caption{Natural and robust accuracy of retrained classifiers on ImageNet. Feature extractors are pretrained robust models trained with $\ell_2$-norm. Robust accuracy is evaluated against adversarial attacks crafted with 20 PGD steps, attack step size 1 and $\epsilon_{te}=\epsilon_{tr}$.}
\label{tab:clf_imagenet}
\vskip 0.15in
\begin{center}
\begin{footnotesize}
\begin{sc}
\begin{tabular}{lllrrr}
\toprule
model & metric & classifier & $\epsilon_{tr}=0.5$ &  $\epsilon_{tr}=3$ &  $\epsilon_{tr}=5$ \\
\midrule
\rn{18} & natural accuracy & original & 65.48 & 53.12 & 45.59 \\
        &                  & retrained & 65.85 & 55.07 & 49.18 \\
        &                  &           & (+0.37) & (+1.95) & (+3.59) \\
        & robust accuracy  & original & 55.15 & 31.05 & 21.85 \\
        &                  & retrained & 54.20   & 27.17   & 16.93 \\
        &                  &           & (-0.95) & (-3.88) & (-4.92) \\
\midrule
\rn{50} & natural accuracy & original  & 73.16 & 62.83 & 56.13 \\
        &                  & retrained & 73.19   & 64.06   & 58.70 \\
        &                  &           & (+0.03) & (+1.23) & (+2.57) \\
        & robust accuracy  & original  & 63.41 & 38.94 & 27.78 \\
        &                  & retrained & 62.31   & 34.53   & 22.10 \\
        &                  &           & (-1.10) & (-4.41) & (-5.68) \\
\midrule
\wrn{50}{2} & natural accuracy & original  & 75.11 & 66.90 & 60.94 \\
            &                  & retrained & 74.87   & 67.48   & 62.73 \\
            &                  &           & (-0.24) & (+0.58) & (+1.79) \\
            & robust accuracy  & original  & 65.85 & 41.70 & 30.61 \\
            &                  & retrained & 64.85   & 37.69   & 25.28 \\
            &                  &           & (-1.00) & (-4.01) & (-5.33) \\
\bottomrule
\end{tabular}
\end{sc}
\end{footnotesize}
\end{center}
\vskip -0.1in
\end{table}

\subsection{Color Bias}
\label{subsec:exp_color_bias}
We evaluate natural accuracy on the extreme case of pixel-averaged images, as discussed in \Cref{subsec:methods_color_bias}.

\begin{table}
\caption{Natural accuracy on pixel-averaged images for CIFAR-10 and CIFAR-100. Robust models are trained with $\ell_2$-norm.}
\label{tab:avg_pixels_cifar}
\vskip 0.15in
\begin{center}
\begin{footnotesize}
\begin{sc}
\begin{tabular}{lcccc}
\toprule
model & $\epsilon_{tr}=0$ &  $\epsilon_{tr}=0.5$ & $\epsilon_{tr}=1$ & $\epsilon_{tr}=2$ \\
\midrule
CIFAR-10 \rn{18} & 8.00 & 16.66 & 15.57 & 17.57 \\
CIFAR-100 \rn{18} & 1.08 & 1.18 & 1.40 & 2.76 \\
\bottomrule
\end{tabular}
\end{sc}
\end{footnotesize}
\end{center}
\vskip -0.1in
\end{table}

\begin{table}
\caption{Natural accuracy on pixel-averaged images for ImageNet. Robust models are trained with $\ell_2$-norm.}
\label{tab:avg_pixels_imagenet_l2}
\vskip 0.15in
\begin{center}
\begin{footnotesize}
\begin{sc}
\begin{tabular}{lrrrr}
\toprule
model &  $\epsilon_{tr}=0$ &  $\epsilon_{tr}=0.5$ &  $\epsilon_{tr}=3$ &  $\epsilon_{tr}=5$ \\
\midrule
\rn{18} & 0.292 & 0.302 & 0.346 & 0.456 \\
\rn{50} & 0.264 & 0.340 & 0.370 & 0.428 \\
\wrn{50}{2} & 0.204 & 0.318 & 0.398 & 0.420 \\
\wrn{50}{4} & 0.184 & 0.312 & 0.446 & 0.404 \\
\vgg & 0.298 & - & 0.390 & - \\
\bottomrule
\end{tabular}
\end{sc}
\end{footnotesize}
\end{center}
\vskip -0.1in
\end{table}

We observe from \Cref{tab:avg_pixels_cifar} that, for CIFAR-10 and CIFAR-100, robust models can achieve better performance than natural models - and better than random guessing - on images that contains no information other than the average color. This trend is confirmed on ImageNet and across different architectures (\Cref{tab:avg_pixels_imagenet_l2}). The second experiment is aimed at highlighting potential failure modes related to the color bias confirmed above. In \Cref{fig:contour_l2} is shown the average (computed over different colors) drop in natural accuracy with respect to clean images when a colored contour is added. Performance of robust models are remarkably impaired, even with relatively thin contours. For example, a 5-pixels contour for ImageNet images amounts for less than $9\%$ of the entire image, yet causes a drop in accuracy that is more than doubled compared to natural models. Besides the larger average drop, adversarially-trained models also present higher variance across different colors, a clear sign of instability under color-based transformations. The results discussed above prove that robust models, while more `stable' than natural models from the perspective of adversarial noise, relies on a delicate balance based on summary statistics of the dataset, as is color. Once a perturbation in this sense is introduced, their stability is compromised and performance decrease catastrophically.

\renewcommand{\fsize}{0.24}
\begin{figure*}
    \centering
\includegraphics[width=\fsize\textwidth]{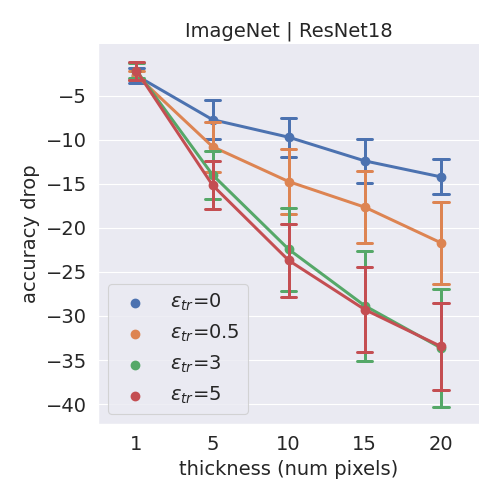}
\includegraphics[width=\fsize\textwidth]{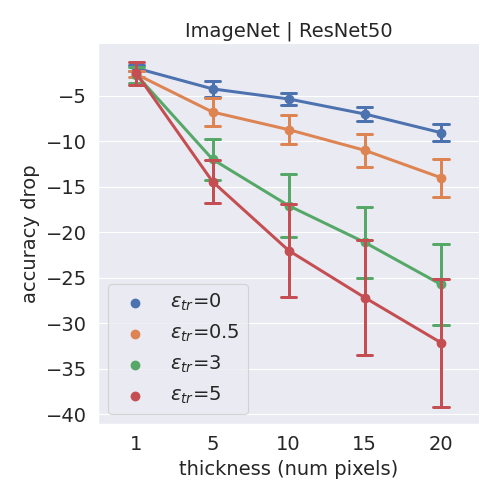}
\includegraphics[width=\fsize\textwidth]{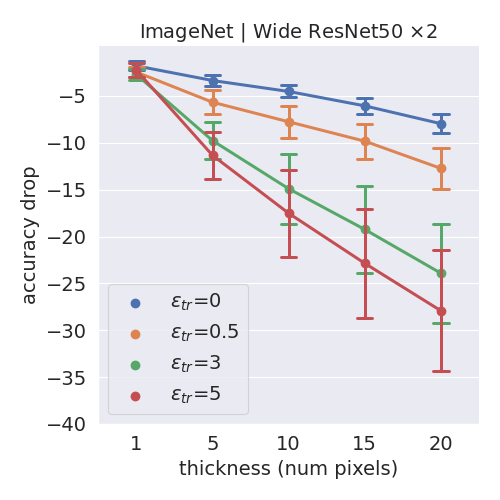}
\includegraphics[width=\fsize\textwidth]{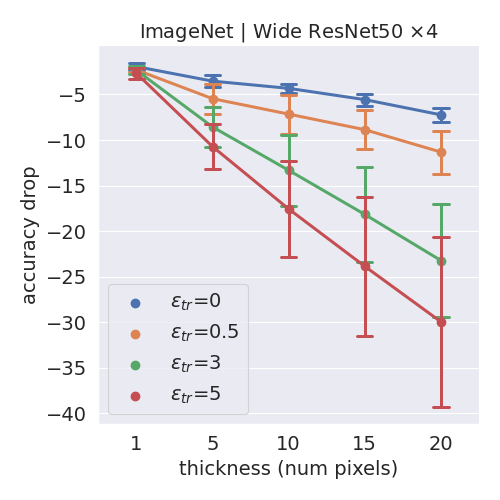}
\includegraphics[width=\fsize\textwidth]{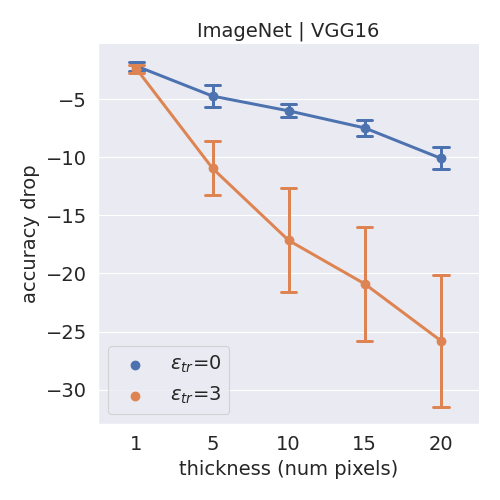}
\includegraphics[width=\fsize\textwidth]{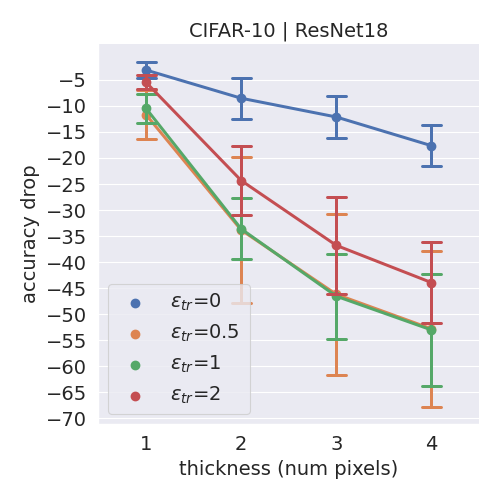}
\includegraphics[width=\fsize\textwidth]{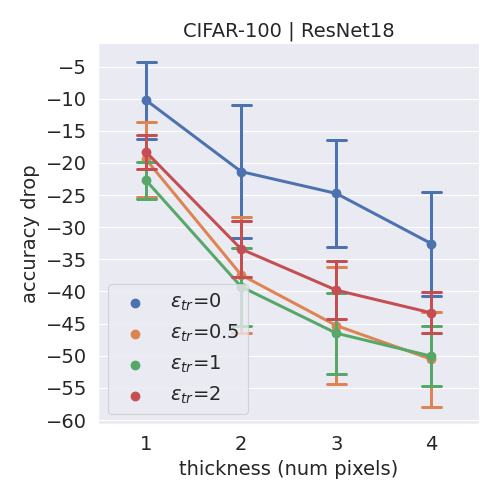}
    \caption{Colored contours: natural accuracy drop with respect to original test set. Circles represent the average accuracy drop over different colors (white, red, green, blue), error bars indicate the corresponding standard deviation. Robust models are trained with $\ell_2$-norm.}
    \label{fig:contour_l2}
\end{figure*}

\section{Discussion}
\label{sec:discussion}
Experiments on always densely active feature maps and feature maps redundancy provide evidence that the effective capacity of robust models is reduced compared to natural models. As an effect of the reduced capacity of hidden layers, robust models struggle in modeling complex concepts and thus resort to simple features such as color. The analysis of the dynamics of always active feature maps and feature maps redundancy during training and their correlation with the natural/adversarial loss could give some hints in this direction.
Orthogonal to the above considerations, experiments on retrained classifiers (on natural examples) shows that the remaining capacity is not fully exploited for the purposes of the learning task - robust classification. In fact, once acknowledged the presence of more predictive, yet less robust combinations of features, a natural question arises: why are these features present if they are not exploited by the classifier? One hypothesis is that they are an artifact of the training process. At the initial stage of the training, a weak attacker might not be able to provide informative adversarial examples and the learning process may be irreversibly biased towards solutions that are not optimal for robust classification. Formulating an exhaustive answer to this question could be an interesting avenue for future research. Together, these facts provide an argument towards explaining (at least partially) the accuracy drop in adversarially-trained CNNs. As discussed in \cite{yang2020closer}, we hypothesize that CNN architectures suffer from structural limitations that prevent them from achieving high robust accuracy without compromising natural accuracy. We will investigate the properties of other architectures - such as Visual Transformers \cite{dosovitskiy2020image} - in future works to test our hypothesis. We will also assess the impact of the training paradigm under the same architecture. To this end, the analysis of multitask learning \cite{ruder2017overview, mao2020multitask} and self-supervised learning \cite{chen2020simple, he2020momentum, kolesnikov2019revisiting}  represents a good starting point.

\section{Conclusion}
\label{sec:conclusion}
In this work, we highlight previously unnoticed properties of adversarially-trained Convolutional Neural Networks, opening the door to novel directions for the understanding of Adversarial Training. Specifically, we find that feature maps in robust models are more densely activated and more redundant than in natural models. These peculiar attributes suggest a close connection between the density of activations, the redundancy of feature maps and the level of robustness in adversarially-trained models. We also challenge some common beliefs about robust models and point to issues that should be carefully addressed. Notably, we show that the color bias of robust models could significantly worsen natural accuracy under semantic-preserving transformations and that their latent space is not inherently robust, contrary to common belief. Taken together, our analyses on the characteristics of feature maps call for a deeper analysis of existing models and offer insights to unveil the inner workings of robust models. Finally, our experiments on color bias and latent features suggest adopting a more critic perspective on widely accepted - yet poorly tested - behaviors. In view of this, we promote experiments aimed at exposing potential failure modes that might be harmful in real-world applications.

\clearpage
\bibliographystyle{ieeetr}
\bibliography{biblio}

\newpage
\appendix

\section{Details on models}
\label{app:models}
For CIFAR-10 and CIFAR-100, all models are trained for 150 epochs, batch size 128, weight decay $5 \cdot 10^{-4}$, initial learning rate $\eta=0.1$ and a drop of $\eta$ by a factor $10$ every 50 epochs. For robust models, adversarial examples are crafted with $7$ PGD steps and attack step size $\alpha=1$.
For ImageNet, we consider the pretrained models from \cite{salman2020adversarially}. Please refer to \cite{salman2020adversarially} for further details.

\section{Omitted results}
\label{app:omitted_results}
\begin{figure}[h]
    \centering
\includegraphics[width=0.6\columnwidth]{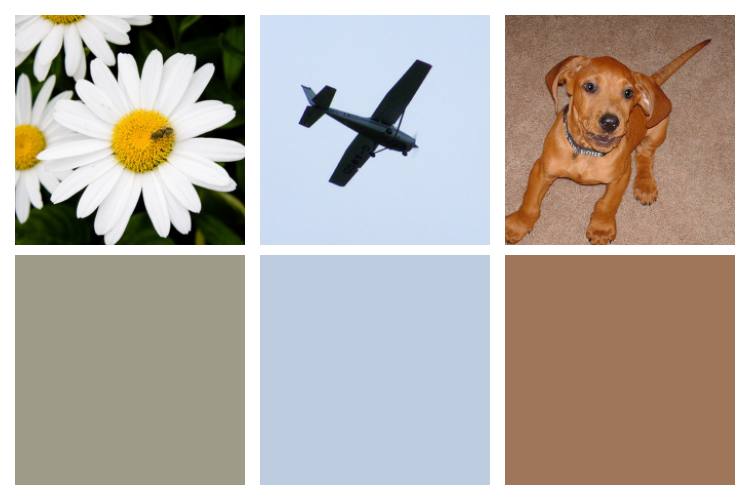}
    \caption{Original images (top) and their pixel-averaged counterparts (bottom) from the ImageNet test set.}
    \label{fig:avg_images_imagenet}
\end{figure}

\begin{figure}[h]
    \centering
\includegraphics[width=0.98\textwidth]{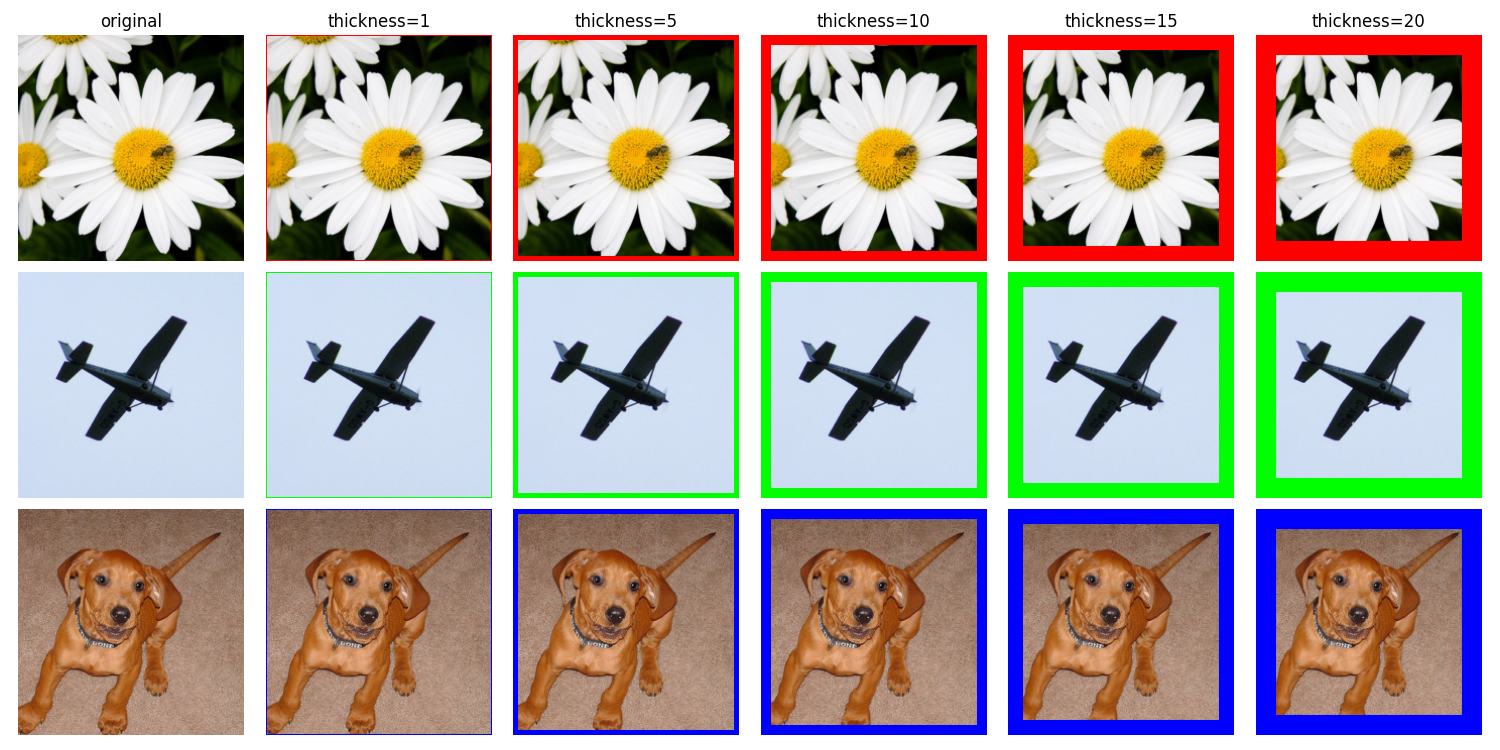}
    \caption{ImageNet images with colored contours.}
    \label{fig:contour_images_imagenet}
\end{figure}

\Cref{fig:num_alw_linf} shows the number of always densely active feature maps for robust models trained with $\ell_{\infty}$-norm. For the same models, \Cref{fig:num_redundant_linf,fig:avg_sim_linf} report the results on feature maps redundancy. In \Cref{tab:avg_pixels_imagenet_linf} the natural accuracies on pixel-averaged images are listed, while \Cref{fig:contour_exp_linf} shows the natural accuracy drop when colored contours are applied. Examples of pixel-averaged images and colored contours are given in \Cref{fig:avg_images_imagenet,fig:contour_images_imagenet}, respectively.

\renewcommand{\fsize}{0.32}
\begin{figure}[h]
    \centering
\includegraphics[width=\fsize\textwidth]{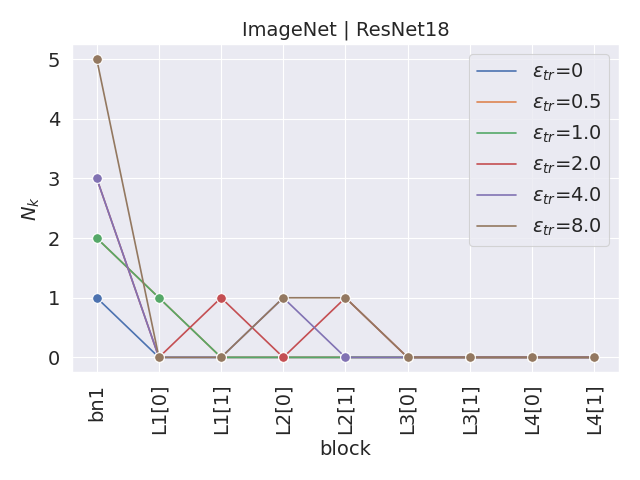}
\includegraphics[width=\fsize\textwidth]{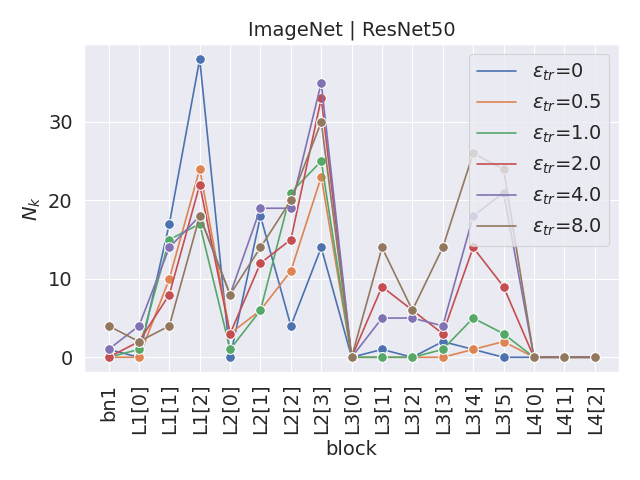}
\includegraphics[width=\fsize\textwidth]{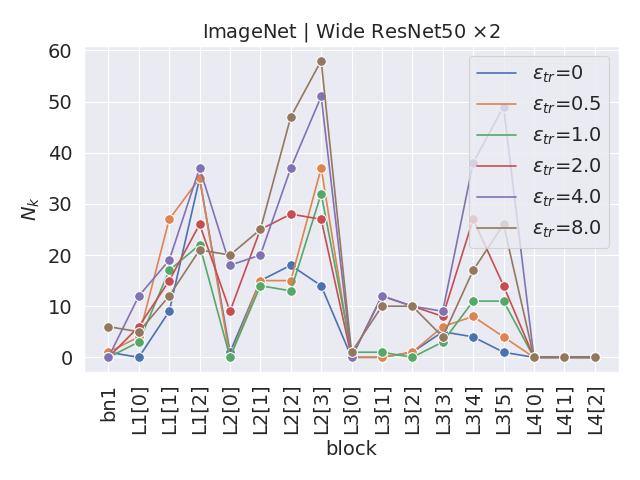}
    \caption{Number of always densely active feature maps (at level $\tau_{den}=0.95$). Robust models are trained with $\ell_{\infty}$-norm.}
    \label{fig:num_alw_linf}
\end{figure}

\renewcommand{\fsize}{0.32}
\begin{figure}[h]
    \centering
\includegraphics[width=\fsize\textwidth]{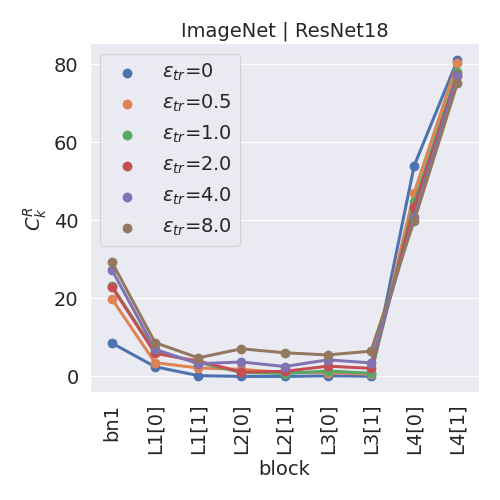}
\includegraphics[width=\fsize\textwidth]{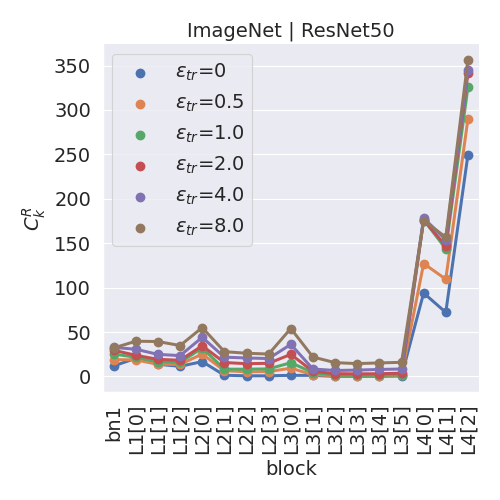}
\includegraphics[width=\fsize\textwidth]{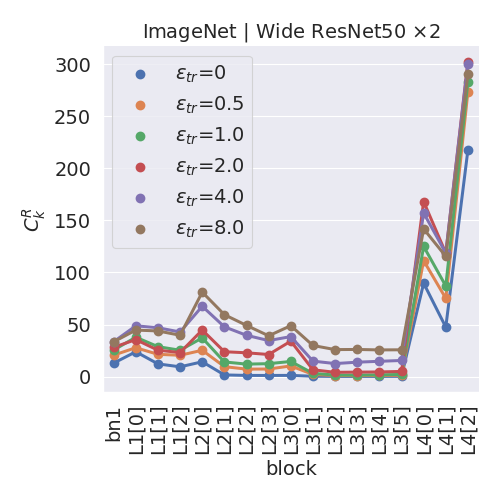}
    \caption{Number of redundant feature maps. Robust models are trained $\ell_{\infty}$-norm. For ImageNet, Values are averaged over 5000 images randomly sampled from the test set.}
    \label{fig:num_redundant_linf}
\end{figure}

\renewcommand{\fsize}{0.32}
\begin{figure}[h]
    \centering
\includegraphics[width=\fsize\textwidth]{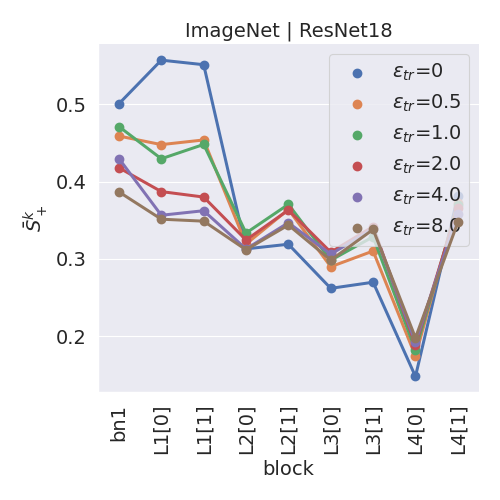}
\includegraphics[width=\fsize\textwidth]{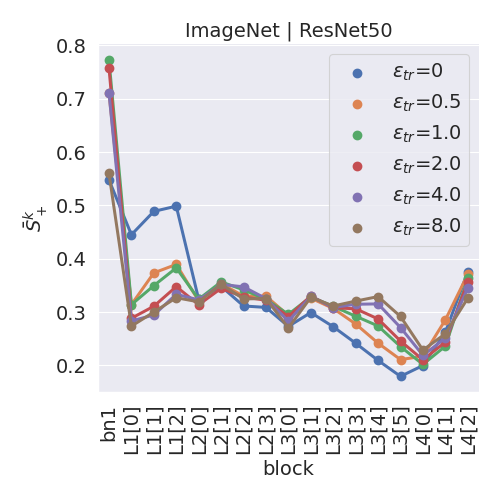}
\includegraphics[width=\fsize\textwidth]{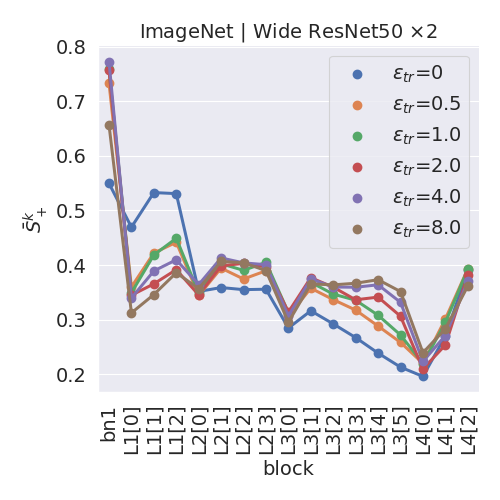}
    \caption{Average cosine similarity. Robust models are trained with $\ell_{\infty}$-norm. For ImageNet, Values are averaged over 5000 images randomly sampled from the test set.}
    \label{fig:avg_sim_linf}
\end{figure}

\begin{table}[h]
\caption{Natural accuracy on pixel-averaged images for ImageNet. Robust models are trained with $\ell_{\infty}$-norm.}
\label{tab:avg_pixels_imagenet_linf}
\vskip 0.15in
\begin{center}
\begin{footnotesize}
\begin{sc}
\begin{tabular}{lrrrrrr}
\toprule
model &  $\epsilon_{tr}=0$ &  $\epsilon_{tr}=0.5$ &  $\epsilon_{tr}=1.0$ &  $\epsilon_{tr}=2.0$ &  $\epsilon_{tr}=4.0$ &  $\epsilon_{tr}=8.0$ \\
\midrule
\rn{18} & 0.292 & 0.358 & 0.362 & 0.406 & 0.410 & 0.436 \\
\rn{50} & 0.264 & 0.352 & 0.322 & 0.314 & 0.364 & 0.340 \\
\wrn{50}{2} & 0.204 & 0.342 & 0.326 & 0.332 & 0.390 & 0.300 \\
\bottomrule
\end{tabular}
\end{sc}
\end{footnotesize}
\end{center}
\vskip -0.1in
\end{table}

\renewcommand{\fsize}{0.32}
\begin{figure}[h]
    \centering
\includegraphics[width=\fsize\textwidth]{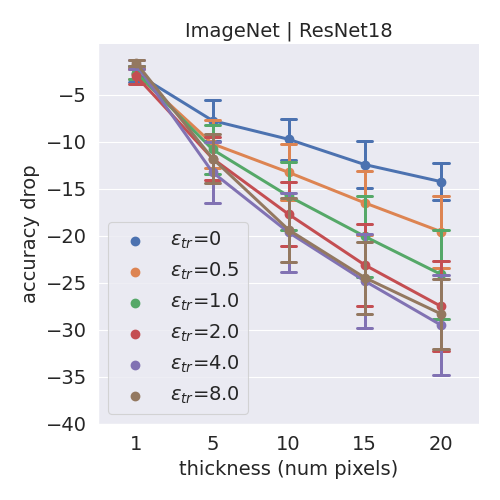}
\includegraphics[width=\fsize\textwidth]{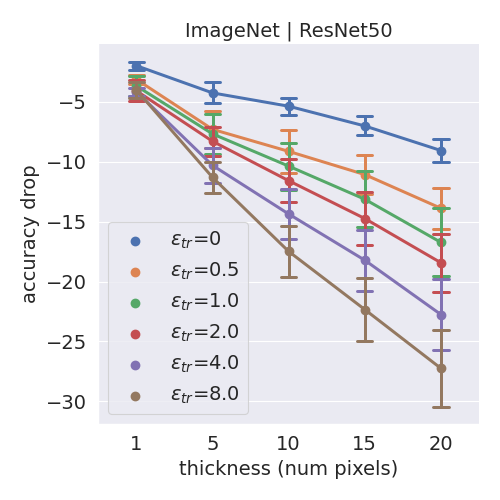}
\includegraphics[width=\fsize\textwidth]{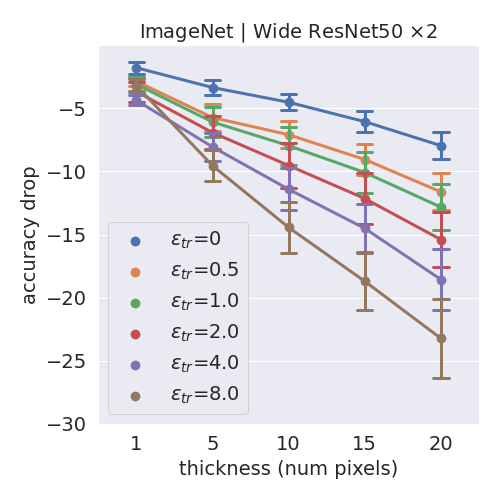}
    \caption{Colored contours: natural accuracy drop with respect to original test set on ImageNet. Circles represent the average accuracy drop over different colors (white, red, green, blue) and error bars indicate the corresponding standard deviation. Robust models are trained with $\ell_{\infty}$-norm.}
    \label{fig:contour_exp_linf}
\end{figure}

\begin{figure}[h]
    \centering
\includegraphics[width=0.98\textwidth]{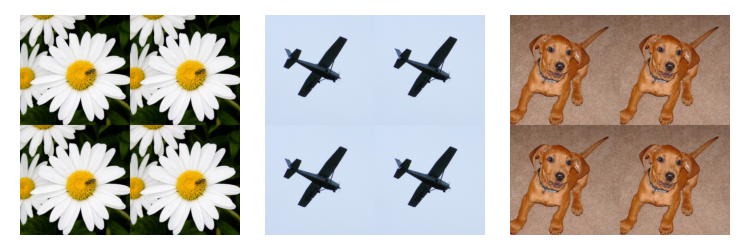}
    \caption{ImageNet images with $4 \times 4$ grid transformation. Each of the $4$ copies has dimensions $224 \times 224$, as the original image. The transformed images have dimensions $448 \times 448$.}
    \label{fig:grid_images_imagenet}
\end{figure}

\clearpage
We demonstrate that the larger accuracy drop of robust models - if compared to natural models - is not limited to the colored contours experiments discussed in \Cref{subsec:exp_color_bias}. We consider a content-preserving transformation in which the original image is replicated and re-arranged in a $4 \times 4$ grid, where each of the $4$ copies preserves the original size (see \Cref{fig:grid_images_imagenet}). Results in \Cref{fig:grid_exp_imagenet} show that robust models are more sensitive than natural models to the grid transformation. This may suggest that robust models are more biased towards the center of the image. Indeed, under the $4 \times 4$ grid transformation, the main object is not positioned near the center of the image - while in the original image it generally is - and models affected by position bias may exhibit a larger accuracy drop. Notice that a recent analysis of pretrained models proved that absolute position information can be implicitly encoded in CNNs \cite{islam2020much}.

\renewcommand{\fsize}{0.32}
\begin{figure}[h]
    \centering
\includegraphics[width=\fsize\textwidth]{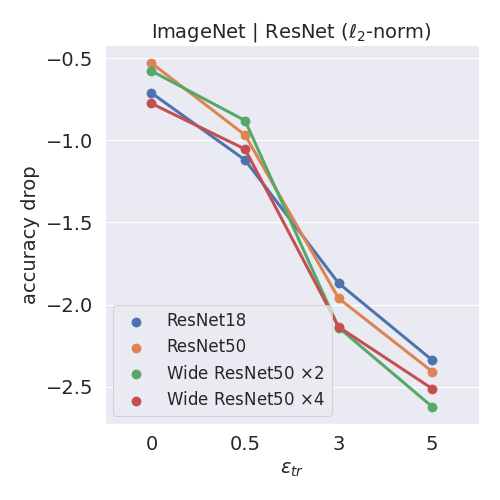}
\includegraphics[width=\fsize\textwidth]{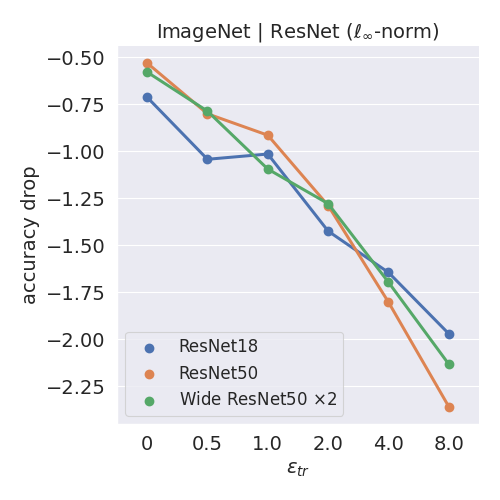}
\includegraphics[width=\fsize\textwidth]{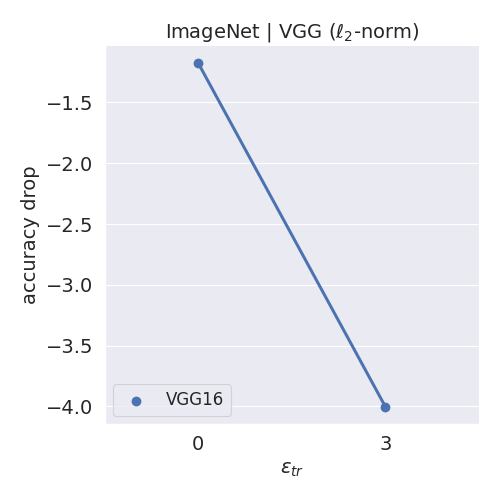}
    \caption{$4 \times 4$ grid: natural accuracy drop with respect to original test set on ImageNet.}
    \label{fig:grid_exp_imagenet}
\end{figure}

\end{document}